%% file: egbib.tex
\documentclass[runningheads]{llncs}
\usepackage{graphicx}
\usepackage{tikz}
\usepackage{comment}
\usepackage{amsmath,amssymb} % define this before the line numbering.
\usepackage{color}

\usepackage{graphicx}
\usepackage{bm}
\usepackage{amsmath}
\usepackage{amssymb}
\usepackage{booktabs}
\usepackage{multirow}
\usepackage{pifont}
\usepackage{hyperref}       % hyperlinks
\hypersetup{
    colorlinks = true
}
\newcommand{\xmark}{\ding{55}}
\usepackage{enumitem}
\usepackage{xcolor}

\usepackage[accsupp]{axessibility}  % Improves PDF readability for those with disabilities.
\definecolor{blue1}{RGB}{43,200,208}

\begin{document}
% \linenumbers
\pagestyle{headings}
\mainmatter
\def\ECCVSubNumber{1693}  % Insert your submission number here
%\ouradd{IncCLIP:}
\title{Generative Negative Text Replay for Continual Vision-Language Pretraining} % Replace with your title
% \ouradd{Reprensentation Learning}  \ourdelete{Pretraining}
% \begin{comment}
% \titlerunning{ECCV-22 submission ID \ECCVSubNumber} 
% \authorrunning{ECCV-22 submission ID \ECCVSubNumber} 
% \author{Anonymous ECCV submission}
% \institute{Paper ID \ECCVSubNumber}
% \end{comment}
%******************

% \begin{comment}
\titlerunning{IncCLIP}
%This work was supported by Shanghai Science and Technology Program 21010502700, and ShanghaiTech and United Imaging Intelligence Joint Lab.
\author{Shipeng Yan\inst{1,2,3} \and
Lanqing Hong\inst{4} \and
Hang Xu\inst{4} \and
Jianhua Han\inst{4} \and
Tinne Tuytelaars\inst{5} \and
Zhenguo Li\inst{4} \and
Xuming He\inst{1,6}
}
%
%\thanks{}

\authorrunning{S. Yan et al.}

\institute{$^\text{1 }$ ShanghaiTech University $^\text{2}$ Shanghai Institute of Microsystem and Information Technology, Chinese Academy of Sciences $^\text{3}$ University of Chinese Academy of Sciences $^\text{4}$ Huawei Noah’s Ark Lab $^\text{5}$~KU Leuven $^\text{6}$~
Shanghai Engineering Research Center of Intelligent Vision and Imaging \\
\email{yanshp@shanghaitech.edu.cn}, \email{\{hanjianhua4, honglanqing, xu.hang, xuchunjing, li.zhenguo\}@huawei.com}, \email{tinne.tuytelaars@kuleuven.be}, \email{hexm@shanghaitech.edu.cn}}
% \end{comment}
%******************
\maketitle

\begin{abstract}
Vision-language pre-training (VLP) has attracted increasing attention recently.
With a large amount of image-text pairs, VLP models trained with contrastive loss have achieved impressive performance in various tasks, especially the zero-shot generalization on downstream datasets.
In practical applications, however, massive data are usually collected in a streaming fashion, requiring VLP models to continuously integrate novel knowledge from incoming data and retain learned knowledge.
In this work, we focus on learning a VLP model with sequential chunks of image-text pair data.
To tackle the catastrophic forgetting issue in this multi-modal continual learning setting, we first introduce pseudo text replay that generates hard negative texts conditioned on the training images in memory, which not only better preserves learned knowledge but also improves the diversity of negative samples in the contrastive loss.
Moreover, we propose multi-modal knowledge distillation between images and texts to align the instance-wise prediction between old and new models.
We incrementally pre-train our model on both the instance and class incremental splits of the Conceptual Caption dataset, and evaluate the model on zero-shot image classification and image-text retrieval tasks.
Our method consistently outperforms the existing baselines with a large margin, which demonstrates its superiority.
Notably, we realize an average performance boost of $4.60\%$ on image-classification downstream datasets for the class incremental split.
% \dots
\end{abstract}

\input{./data/introduction.tex}
\input{data/relatedwork.tex}
\input{data/methods.tex}
\input{data/exps.tex}
\input{data/discussion.tex}
\input{data/conclusion.tex}

\clearpage
\bibliographystyle{splncs04}
\bibliography{egbib}
\end{document}

%% file: data/introduction.tex
\section{Introduction}\label{sec:intro}

\begin{figure}[t]
  \centering
%   \fbox{\rule{0pt}{2in} \rule{0.9\linewidth}{0pt}}
   \includegraphics[width=0.65\linewidth]{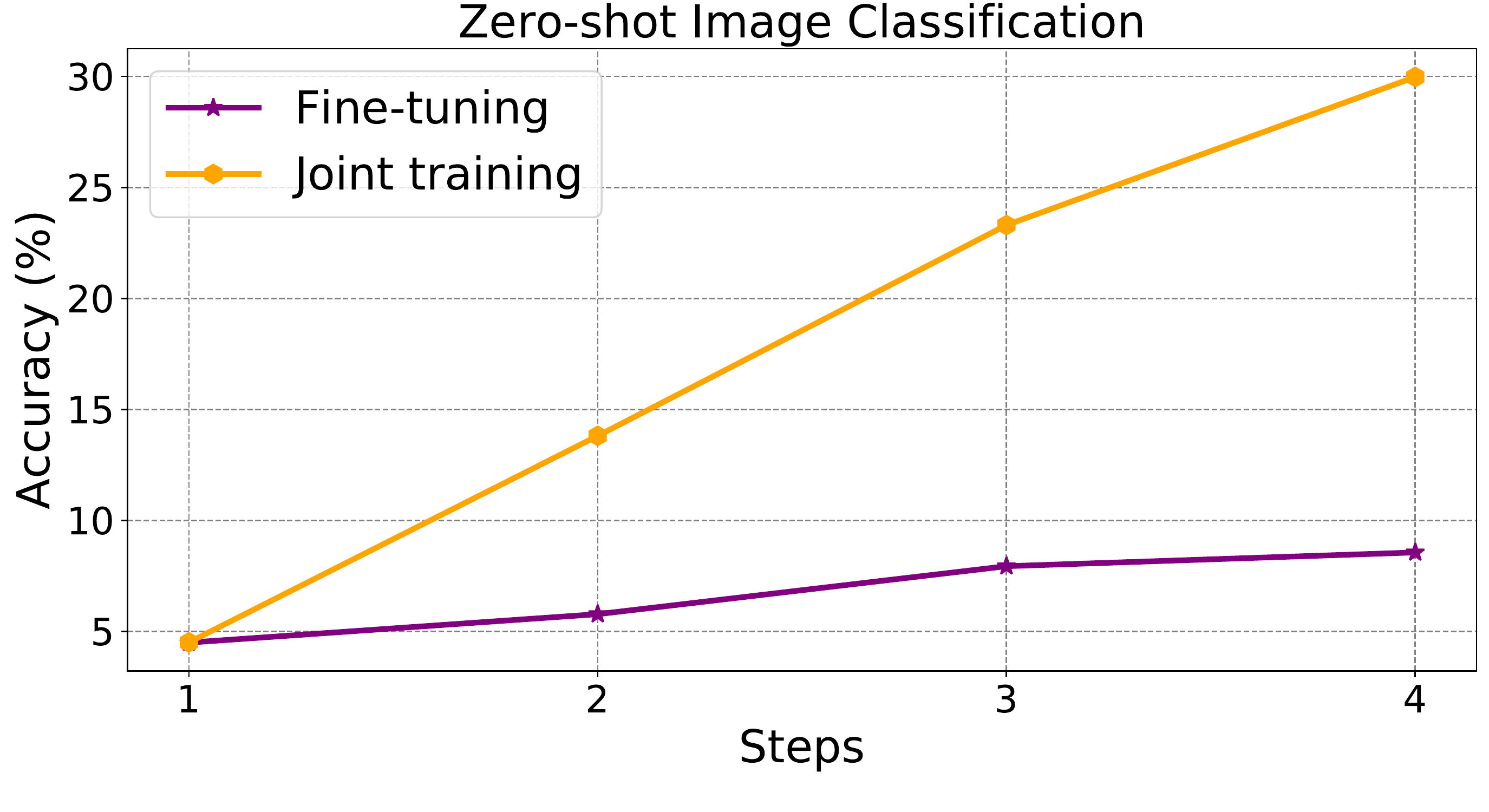}
   \caption{\textbf{Illustration of continual vision-and-language pre-training on CC2M dataset.}
   The dataset is split into four data chunks with 0.5 million image-text pairs in each chunk. In fine-tuning, a new data chunk arrives at each step to update the pre-training model without previous data.
   In joint training, all data are accessible and shuffled during the whole training process.
   The figure shows significant performance gap between the fine-tuning and the joint training strategies on downstream zero-shot classification task.}
   \label{fig:performance_gap}
\end{figure}

% VLP
% Zero-shot generalization is important
Vision-and-language pre-training (VLP)~\cite{pmlr-v139-kim21k,radford2021learning} seeks to learn a generalizable multi-modal representations from large-scale image-text data.
% Collect massive noisy data from websites
Recently, VLP models, such as CLIP~\cite{radford2021learning}, have demonstrated promising performance especially on the zero-shot generalization for a variety of downstream vision-language tasks including zero-shot image classification and image-text retrieval~\cite{han2021contrastive}. % and segmentation~\cite{li2020consistent}.
However, training a CLIP model typically requires a large amount of image-text pairs (400 million),  %collected from a variety of publicly available sources on the Internet. This 
which is particularly burdensome for the traditional off-line training strategy as all the data need to be available during the entire training process. 
%In this case, it is computational burdensome to pre-train the model in a traditional joint-training manner, where all data needs to be accessible during the whole training process.
% data for VLP task often comes in streaming way.
Moreover, for practical applications, it is critical for a VLP model to continuously integrate novel knowledge in a dynamic environment, e.g., from streaming data crawled from the Internet. % every day.
% Large gap compared to the bound
On the other hand, as shown in Figure~\ref{fig:performance_gap}, a naive fine-tuning strategy for VLP using only the incoming new data suffers from a large performance degradation compared to the off-line training strategy.
% Goal; Rare works focus on this important problem.
Consequently, it is essential to address this continual learning problem for large-scale VLPs, a topic that has received little attention in the past.   
%In this work, our goal is to leverage the streaming image-text pairs to continuously improve our VLP model, 
% Real applications? 

% dataset, the meaning of step, how many steps, the setting.

% Challenges: forgetting and preserving features
%In continual learning, the stability-plasticity dilemma~\cite{grossberg2013adaptive} is the primary issue, where a model is prone to forgetting the previously learned knowledge when adapting to new data.
Traditional continual learning methods mostly focus on the issue of stability-plasticity dilemma~\cite{grossberg2013adaptive,wang2021ordisco}, where a model is prone to forgetting the previously learned knowledge when adapting to new data. As such, much effort has been devoted to preserving discriminative features for the known classes~\cite{delange2021continual}.
For the VLP models, however, the pre-trained multi-modal representations need to be transferred to unseen downstream tasks. 
In this case, what knowledge should be retained and how it is preserved is less obvious. % seldom investigated.
In addition, the forgetting problem in continual multi-modal learning involves the representation of visual and language inputs, as well as the multi-modal correspondence between the two modalities, which further complicates the problem.

% Drawbacks of related works
Most state-of-the-art approaches~\cite{Liu2020MANets} in traditional class incremental setting rely on the memory replay of representative old training samples~\cite{delange2021continual}. Nevertheless, unlike the supervised class incremental learning~\cite{castro2018end} which typically trains a model to classify on a closed set of categories, VLP models learn an open set of visual concepts from natural language descriptions.
Consequently, the majority of existing memory-based continual learning strategies would be rendered ineffective. In particular, those methods~\cite{zhao2020maintaining,wu2019large} aiming to alleviate class imbalance are not applicable to the continual learning of VLP models as there are no class-level supervision for VLP.   
% Besides, the logit-based distillation methods~\cite{rebuffi2017icarl} operate on explicit label space, which does not exist in VLP tasks. %require to access both the old and new classes in advance, which is not realistic in VLP tasks. 
In addition, the feature-based distillation methods~\cite{douillard2020podnet} are usually designed for uni-modal CNNs and rarely take into account the characteristics of transformer-based multi-modal pre-training, and the model inversion~\cite{yin2020dreaming} aims to generate images as positive examples for certain classes using a frozen copy of trained models, which remains challenging for high-resolution images and may introduce biases to the generated training data~\cite{wang2020cnn}.

%To tackle the stability-plasticity dilemma, memory replay of representative old training samples has been shown as an effective strategy~\cite{delange2021continual}, which achieves state-of-the-art (SOTA) performance in traditional class incremental setting and has been adopted by a large majority of continual learning works~\cite{Liu2020MANets}.
% Our method follows replay-based but current methods are not applicable

% Do not have parametric classifier.
%Thus the approaches~\cite{zhao2020maintaining}~\cite{wu2019large} aiming to alleviate class imbalance are not applicable to the continual learning of VLP models.
% closed world

% Moreover, there are biases in the generated images that are easily caught by the discriminative model which eventually leads the model losing its ability to generalize~\cite{wang2020cnn}.
% Replay-based methods~\cite{} store part of historical data for rehearsal and have proven to be effective to overcome forgetting in a class incremental setting.
% It has been adopted by a large majority of continual learning works.
% Our method also follows replay-based?
% Continual vision-language pre-training (CL-VLP) is a fresh and important research problem, and lots of known continual learning methods are not applicable to continual vision-language pre-training.

% Our Approach
%we aim to tackle the aforementioned stability-plasticity trade-off in continual multi-modal pre-training setting.
In this work, we propose a novel replay-based continual learning framework, named as IncCLIP, to address the above challenges for the VLP tasks.
To this end, based on CLIP~\cite{radford2021learning}, we introduce a conditional data generation process that extracts the `dark' knowledge from the previous-step model in a form of pseudo texts for replay, and adopt multi-modal knowledge distillation loss to further overcome catastrophic forgetting.
Specifically, our model architecture is a two-stream encoder composed of separate visual and text encoders.
At each incremental stage, in order to learn generic and transferable visual-linguistic representation, we adopt a contrastive loss that requires the model to predict the pairing between images and texts.
Given the importance of negative instances in contrastive learning, we introduce a pseudo text generation technique via model inversion~\cite{yin2020dreaming} to augment the data memory with informative data and replay the generated texts as negative examples.
Moreover, to alleviate catastrophic forgetting, we design a knowledge distillation loss to minimize the output discrepancy between current and the previous-step model, which preserves the knowledge on cross-modal correspondence.
It is also worth noting that after the incremental training of each step, we adopt reservoir sampling to update the memory for the rehearsal in the next step.

To validate our method, we perform continual model pre-training on an instance incremental and a class incremental split for the Conceptual Caption dataset~\cite{sharma2018conceptual}, which simulates two different real scenarios.
%\ourdelete{on two large-scale datasets including CC12M dataset and its subset CC2M, respectively}.
We then evaluate our model on two downstream tasks: zero-shot image classification and zero-shot image-text retrieval. 
The experimental results demonstrate the superiority of our approach, which is then further analyzed via the detailed ablation study. Notably, we outperform previous approaches by 4.6$\%$ in accuracy on the downstream image classification task with four-step class incremental split. %\ouradd{where the image class disribution shifts over time}. 
% Contributions
In summary, the main contributions of our work are three-fold:
\begin{itemize}%[itemsep=-1mm,topsep=1mm]
    \item To our best knowledge, this is the first work to tackle the problem of continual vision-language pre-training with streaming image-text pairs.
    \item To achieve better stability-plasticity trade-off, we propose the IncCLIP framework to augment the contrastive learning with the negative pseudo texts and adopt a multi-modal knowledge distillation loss between images and texts to preserve the learned cross-modal correspondence.
    \item Our proposed method consistently outperforms previous CL baselines such as UCIR in standard continual vision-language pre-training benchmarks.
\end{itemize}

%% file: data/relatedwork.tex
\section{Related Works}\label{sec:related_works}

\subsubsection{Vision-Language Pre-training}
%Pre-training for vision-language (VL) tasks has been proven to be successful in learning transferable joint image-text embeddings from large-scale image-text pairs.
Vision-language pre-training learns transferable joint image-text embeddings from large-scale image-text pairs, which has been shown to be effective for a variety of vision-language (VL) tasks~\cite{ALBEF,liu2021kd,cho2021unifying}.
% Model architecture , 
The majority of existing works fall into two categories based on model architectures: single-stream and dual-stream models.
Single-stream models~\cite{chen2020uniter,pmlr-v139-kim21k,liu2021kd} introduce powerful encoders such as Transformer~\cite{vaswani2017attention} to model the cross-modal interaction between image and text. As such, they perform well on VL tasks like VQA~\cite{antol2015vqa}, which requires complex reasoning between image and text. However, they typically use an external object detector to generate visual region descriptions as the input to the multi-modal encoder~\cite{cho2021unifying,chen2020uniter}, which can be computationally expensive.
%By incorporating an external object detector, it typically generates visual region 
Moreover, it is difficult to apply them to certain VL tasks such as image-text retrieval, which requires feeding all potential image-text pairs into the multi-modal encoder and hence is inefficient.

On the other hand, dual-stream models~\cite{jia2021scaling,radford2021learning} adopt a dual-encoder architecture to encode images and texts, respectively. 
CLIP~\cite{radford2021learning} and ALIGN~\cite{jia2021scaling} perform pre-training on large-scale noisy data collected from the Internet.
Especially, CLIP provides a flexible zero-shot classifier rather than parametric task-specific classifiers, and demonstrates impressive zero-shot generalization ability on many downstream tasks, such as zero-shot image classification and zero-shot image-text retrieval.
% Efficient on image-text retrieval tasks
% Dual-stream models are computation-efficient for the image-text retrieval task because they only need single inference on each text and image.
It is a significant step towards flexible and practical zero-shot classifiers for computer vision tasks.
% why VLP needs continual
Nevertheless, current dual-stream VLP models are trained in a joint-training manner using data prepared in advance, without the ability to continuously adapt to new data from a dynamic environment.
%  Not only will new objects with new properties emerge, but the way humans talk about objects changes over time.
In this work, we concentrate on the continual vision-language representation learning based on dual-stream models like CLIP.

\subsubsection{Continual Learning}
% To refine.
Continual learning~\cite{zhaomemory,delange2021continual,thrun1998lifelong,xie2022general} aims to integrate novel knowledge in a sequential fashion where old data are usually unavailable. 
Existing literature focuses mostly on supervised continual learning~\cite{yan2021framework,zhao2021video,simon2021learning}, which mainly falls into the following four groups.
The first is the regularization methods~\cite{ebrahimi2019uncertainty,kirkpatrick2017overcoming}, which penalize changes on significant weights of previously learned models.
The second group is the distillation methods~\cite{li2017learning}, which aim at retaining the output of the network on available data.
The third is the structure methods~\cite{yan2021dynamically,serra2018overcoming}, which keep old parameters fixed while growing and allocating weights for learning new data.
The last is the pseudo rehearsal methods~\cite{yin2020dreaming,smith2021always}, which usually train a generative model to generate visual images of previously learned categories and train the classifier with the combination of real data and pseudo data to reduce forgetting.
%CL on VG?
Our method combines the distillation method and the pseudo rehearsal method. However, previous pseudo rehearsal methods adopt either generative adversarial networks or variational auto-encoders, which are not easy to address their data degeneration issue, especially when dealing with complex scenarios such as high-resolution images or images of the similar classes. Our method circumvents this problem by “inverting” the model of last step to synthesize hard negative texts in the token embedding space, since token embedding has much lower dimensions and generating negative data points is easier. This is inspired by DeepInversion~\cite{yin2020dreaming}, but they do not consider generating negative data points and it is designed for the convolution network with Batch Normalization.

Recently, there also have been some efforts~\cite{rao2019continual,hu2021well,cha2021co2l} to explore continual representation learning with unlabeled streaming data. CURL~\cite{rao2019continual} proposes a continual unsupervised representation learning method, which learns a task-specific representation based on a set of share parameters and also trains a generative model to avoid forgetting.  $\text{Co}^2\text{L}$\cite{cha2021co2l} introduces a self-supervised knowledge distillation method for self-supervised continual learning. However, previous methods are designed for continual learning problems with a single modality and do not consider the properties of multi-modal representation learning. To the best of our knowledge, our work is the first to explore continual learning in the self-supervised multi-modality representation learning.

%% file: data/methods.tex
\section{Methods}\label{sec:methods}
\begin{figure*}[t]
  \centering
%   \fbox{\rule{0pt}{2in} \rule{0.9\linewidth}{0pt}}
   \includegraphics[width=\linewidth]{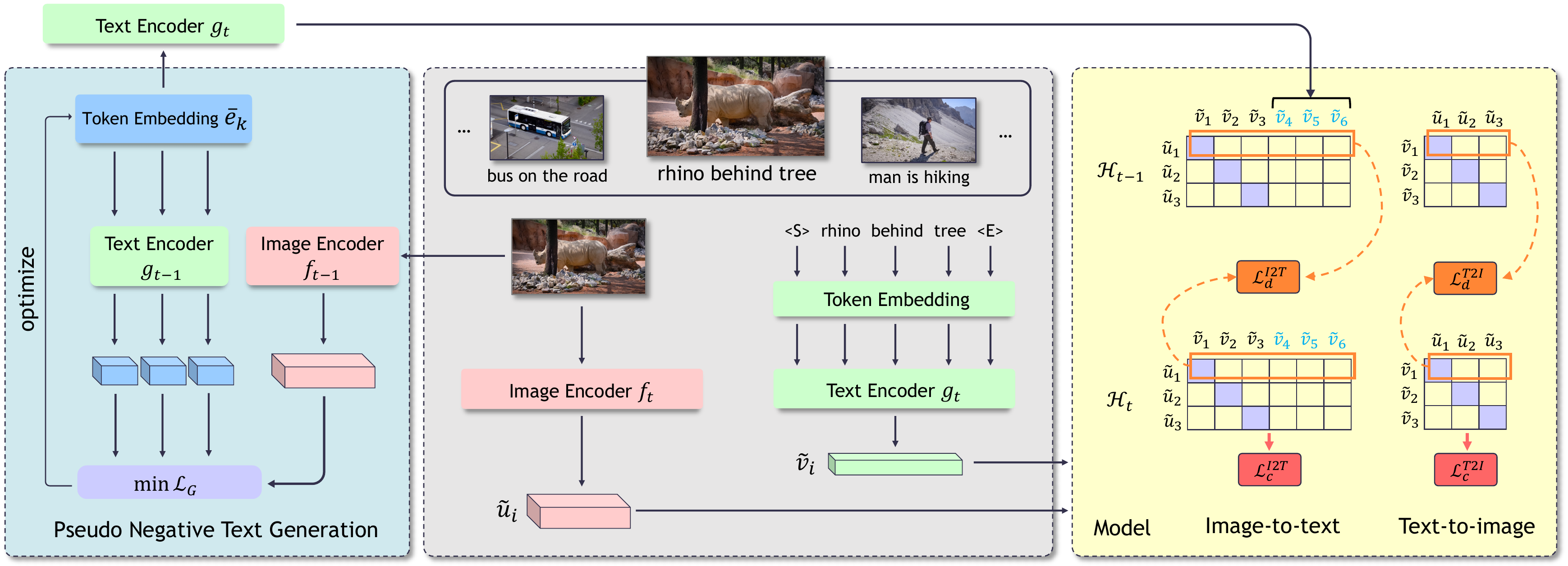}
   \caption{The illustration of our method. 
   The left panel illustrates the pseudo negative text generation process.  We optimize pseudo texts $\bm{\bar{e}}_k$ in the token embedding space by minimizing the loss $\mathcal{L}_G$.
As middle panel shows, for a sampled mini-batch, we first extract the images features $\bm{\tilde{u}}_i$ and language features $\bm{\tilde{v}}_i$ through the corresponding encoder $f_t$ and $g_t$.
   The right panel shows the calculation of training loss. We compute the similarity matrix $S$ between image features $\bm{\tilde{u}}_i$ and language features $\bm{\tilde{v}}_i$. Then we apply the contrastive loss $\mathcal{L}^{\text{I2T}}_c$, $\mathcal{L}^{\text{T2I}}_c$  and distillation loss $\mathcal{L}_d^{\text{I2T}}$,$\mathcal{L}_d^{\text{T2I}}$ from both image to text and text to image.
   Note that $\textcolor{blue1}{\bm{\tilde{v}_i}}$ is the deep feature of the pseudo token embedding $\bm{\bar{e}_i}$, which is used as negative examples in the training loss.}
   \label{fig:method}
\end{figure*}

% Overview
In this section, we describe our approach, as sketched in Figure~\ref{fig:method}, to address the continual vision-language representation learning problem, with the goal of improving stability-plasticity trade-off.
Concretely, we combine contrastive loss and multi-modal knowledge distillation and supplement the training with pseudo texts to enhance the generalization ability for the representation.
Below, we first present the overview of problem setup and model architecture in Sec.~\ref{subsec:setup_arch}, followed by the introduction of text generation in Sec.~\ref{subsec:text_gen}.
Then we detail the training loss in Sec.~\ref{subsec:losses}.

%based on the contrastive loss used in CLIP, we further introduce a replay-based approach that combines the pseudo text replay and multi-modal knowledge distillation to improve the generalization ability for the representation.

\subsection{Problem Setup and Model Architecture}\label{subsec:setup_arch}
% Problem Setup
\subsubsection{Problem Setup} We first introduce the problem setup of \textbf{continual vision-language pre-training}.
During sequential pre-training, the model observes a sequence of data chunks $\mathcal{D}_t$ for each step $t$.
Particularly, the dataset $\mathcal{D}_t=\{(\bm{x}^I_i, \bm{x}^T_i)\}_{i=1}^{|\mathcal{D}_t|}$ is composed of image-text pairs at step $t$, where $\bm{x}^I_i$ means the input image, and $\bm{x}^T_i$ represents the corresponding text describing the image.
We assess the transferability of learned representation to downstream tasks after the training at each chunk.
% Allow to save some data as memory for rehearsal.
In this work, all methods including our method and the comparison methods are based on the rehearsal strategy, which stores a subset of observed data in memory $\mathcal{M}_t=\{(\bm{x}^I_i, \bm{x}^T_i)\}$ for future training.

% Introduce the input
\subsubsection{Model Architecture}

Like CLIP~\cite{radford2021learning}, we adopt a dual-stream encoder structure where the model $\mathcal{H}_t$ with parameters $\theta_t$ has independent visual encoder $f(\cdot)$ and text encoder $g(\cdot)$.
Concretely, given an image $\bm{x}^I_i$ and a text $\bm{x}^T_j$, we first compute the normalized image embedding $\bm{\tilde{u}_i}$ and normalized linguistic embedding $\bm{\tilde{v}_j}$, and then compute the similarity score $s_ij$.

Concretely, to encode the image, we adopt ResNet as visual backbone, which extracts the features $\tilde{\bm{u}}_i=f(\bm{x}^I_i)$ from image $\bm{x}^I_i$, and normalize the feature $\bm{\tilde{u}}_i=\bm{\tilde{u}}_i/\lVert \bm{\tilde{u}}_i \rVert_2$ onto unit-sphere.
To encode the text, we firstly tokenize the sequence into a sequence of word tokens $\bm{e}^T_i$ where we 
use the lower-cased byte pair encoding (BPE)~\cite{sennrich2015neural} with a vocabulary size of 49,408 to tokenize the text.
We adopt Transformer~\cite{vaswani2017attention} as text encoder and  encode these tokens into normalized linguistic embedding $\bm{\tilde{v}}_j = g(\bm{x}^T_j)/\lVert g(\bm{x}^T_j) \rVert_2$.
% For computational efficiency, the max text sequence length is capped at 77.
Finally, we compute the similarity score $s_{ij}=\bm{\tilde{u}}_i^{\top}\bm{\tilde{v}}_j$ between i-th image $\bm{x}^I_i$ and j-th text $\bm{x}^T_j$.

\subsection{Pseudo Negative Text Generation}\label{subsec:text_gen}

In this subsection,  we introduce the pseudo text generation via model inversion to distill the knowledge of last step model $\mathcal{H}_{t-1}$ for solving the catastrophic forgetting.
Due to the distribution shift between synthetic and real examples, the model is frequently biased if regarding them as positive examples~\cite{smith2021always}.
We bypass this issue via generating negative texts which do not guide the model incorrectly.
Having access to informative negative samples is known to be critical for the success of contrastive learning~\cite{he2020momentum}.
Furthermore, it is observed that negative examples, especially hard negative examples, benefit the learning of representation~\cite{robinson2021contrastive}.
Motivated by this, we propose to generate the negative text via model inversion as follows.
% The difficulty of generation
% Image synthesis is more difficult 
% Image synthesis is difficult due to data degeneration issues like mode collapse especially for high-resolution images considering that the optimization space is too large.
% The importance of negative examples in contrastive learning

% Why generate data
% Why choose the approach to generate text(Pros)
% Image/Text Generation is difficult.
% Why generate negative text
% Why generate hard negative text

% How: Ovreview
% We perform the pseudo data generation at the beginning of each step except the initial step.
For each training batch, we perform the pseudo data generation to augment the mini-batch.
Given the discrete nature of tokens, which makes the optimization difficult, we optimize the pseudo text $\bar{\bm{e}}_k$ in the token embedding space to find the hard negative texts with respect to the images $\bm{x}^I_i$.
$\bm{e}_k$ should keep a moderate distance from $\bm{x}^I_i$ as a close distance indicates $\bm{e}_k$ is a positive sample and a remote distance means $\bm{e}_k$ is easy to be distinguished.
%We construct hard negative texts $\bm{e}_k$ by inverting the text encoder in an image-to-text retrieval way.
% Considering the computation efficiency, we generate the hard negative texts using only the currently sampled mini-batch $\{(x^I_i, x^T_i)\}$.
% we don't introduce extra generator.
% Initialization

Specifically, to generate the pseudo texts, we firstly sample a mini-batch $\mathcal{B}=\{(\bm{x}^I_i, \bm{x}^T_i)\}_{i=1}^B$ from memory $\mathcal{M}_t$.
We initialize the token embedding $\bar{\bm{e}}_k=\beta \bm{e}^T_i + (1-\beta) \bm{e}^T_j$ where $\bar{\bm{e}}_k$ are k-th generated text,  $\beta$ is uniformly sampled from the interval $[0,1]$, $i,j$ are indices randomly sampled from 1 to batch size B, and $\bm{e}^T_i$, $\bm{e}^T_j$ are the corresponding token embeddings.
% Why take linear combination -> refer to mixup sth about manifold?
It is worth noting that $\bar{\bm{e}}_k$ takes the value of a real token embedding  as initialization when $\beta=0$ or $\beta=1$.
% Optimization
% Scores
%, which ranges from -1 to 1
Given the current pseudo token embedding $\bar{\bm{e}}_k$, we compute the cosine similarity scores $s_{ik}$ with the images features $\bm{\tilde{u}}_i$.
% Naive
Then we generate the data in the token embedding space which is continuous and easy to optimize compared to discrete tokenization space. %
%which is a continuous space and easy to optimize compared with 
The naive text generation via model inversion is to directly minimize the cosine similarity, which are calculated by the inner product between the generated token sequences $\bar{\bm{e}}_k$ and the sampled image features $\bm{\tilde{u}}_i$.
% Hard negative but not be positive
However, to improve data efficiency, we require the generated token embedding $\bar{\bm{e}}_k$ to be hard negative examples which means they are difficult to be distinguished from positive examples.
To achieve this, we adopt margin loss as follows
\begin{align}
    \mathcal{L}_G = \frac{1}{B} \sum_{i=1}^B \max(0, s_{\text{min}} - s_{ik}) + \max(0, s_{ik} - s_{\text{max}}),
\end{align}
where $s_{\text{min}}$ and $s_{\text{max}}$ are the hyper-parameters representing the minimum and maximum score, respectively.
The maximum score guarantees the generated example to be negative examples.
The minimum score requires the generated examples to be hard negative examples.
We adopt SGD to minimize the loss $\mathcal{L}_G$ with respect to the variable $\bar{\bm{e}}_k$ for a constant number of iterations. 
%TODO: \xh{why this loss works? add one comments}

\subsection{Training Loss}\label{subsec:losses}
% Overview -> Loss detail
%======
% Overview
We now describe the details of our continual pre-training objective.
% Contrastive -> pretraining task； Why use contrastive loss(pros)
For pre-training, we adopt a contrastive learning task to learn a generic and transferable visual-linguistic representation, which has proven to be efficient and effective in previous works~\cite{ALBEF}.
% Knowledge distillation
To preserve the knowledge in the model, we maintain the relation of instances by introducing the knowledge distillation.
% Contrastive learning success key: negative example.(Ref)
% In this work, we generates hard negative text (token embedding) for replay.
With the above generated texts $\bar{\bm{e}}_k$, we use it in both contrastive learning and multi-modal knowledge distillation, which helps the contrastive learning on novel data and improves the regularization of distillation loss.
% Detailedly, Now we have mini-batch (image,text) and text augmentaion e_i as negative examples.

During continual learning, we learn the model on the union of incoming data, memory and pseudo texts.
In detail, we sample a mini-batch of image-text pairs $\{(\bm{x}^I_i, \bm{x}^T_i)\}_{i=1}^B$ from incoming data  $\mathcal{D}_t$ and memory $\mathcal{M}_t$ where $B$ denotes the batch size, and sample a mini-batch of token embedding $\{\bm{e}^T_k\}_{k=1}^{\hat{B}}$ of negative texts with batch size $\hat{B}$.
We denote the text batch size in total as $B_T=B+\hat{B}$.
The linguistic embedding for the sampled texts $\bm{x}^T_i$ and the token embedding $\bm{e}^T_k$ of negative texts are denoted as $\{\bm{\tilde{v}}_j\}_{j=1}^{B_T}$, and the visual embedding are $\{\bm{\tilde{u}}_i\}_{i=1}^B$.
% Illustration of sij? redundant?
We can compute the classification probability $P_{I2T}(y_i|\bm{\tilde{u}}_i, \{\bm{\tilde{v}}_j\}_{j=1}^{B_T})$, $y_i\in \{1,2,\dots,B_T\}$, from image to texts to determine which text corresponds to the given image as follows
\begin{align}
    P_{I2T}(y_i=k|\bm{\tilde{u}}_i, \{\bm{\tilde{v}}_j\}_{j=1}^{B_T};\tau) = \frac{\exp(s_{ik}/\tau)}{\sum_{j=1}^{B_T} \exp(s_{ij}/\tau) },\end{align}
where $\tau$ is the temperature parameter to control the smoothness of the Softmax function, and $P_{I2T}(y_i=k|\bm{\tilde{u}}_i, \{\bm{\tilde{v}}_j\}_{j=1}^{B_T};\tau)$ means the chance of i-th image being paired with k-th text.
Similarly, we can compute the classification probability $P_{T2I}(y_j|\bm{\tilde{v}}_j, \{\bm{\tilde{u}}_i\}_{i=1}^B)$ from text to images, i.e. the chance of j-th text being paired with k-th image, as follows
\begin{align}
    P_{T2I}(y_j=k|\bm{\tilde{v}}_j, \{\bm{\tilde{u}}_i\}_{i=1}^B; \tau) = \frac{\exp(s_{kj}/\tau)}{\sum_{i=1}^B \exp(s_{ij}/\tau)}.
\end{align}
% where $y\in \{1,2,\dots,B\}$ means the image prediction and the probab is most closed related to the given text.
\subsubsection{Contrastive Loss}
We adopt bi-directional contrastive loss to learn generalized image representations from natural language supervision.
We jointly train an image encoder and a text encoder to predict the correct pairings of a batch of image-text pairs.
Concretely, for an image $\bm{x}^I_i$, we regard its corresponding language description $\bm{x}^T_i$ as a positive example whereas the other $B_T-1$ texts are considered negative examples. 
Therefore, the image-to-text loss is as follows 
%(row-wise/column-wise)
\begin{align}
    \mathcal{L}^{I2T}_c = - \frac{1}{B} \sum_{i=1}^{B} \log P_{I2T}(y_i|\bm{\tilde{u}}_i, \{\bm{\tilde{v}}_j\}_{j=1}^{B_T}; \tau),
\end{align}
where the ground truth $y_i \in \{1,2,\dots,B\}$. 
The text-to-image loss $\mathcal{L}^{T2I}_c$ is defined on the texts $\{\bm{x}^T_i\}_{i=1}^B$ in a similar way as follows
\begin{align}
    \mathcal{L}^{T2I}_c = - \frac{1}{B} \sum_{i=1}^B \log P_{T2I}(y_j|\bm{\tilde{v}}_j, \{\bm{\tilde{u}}_i\}_{i=1}^B; \tau),
\end{align}
where we only compute the loss on real texts because the pseudo texts lack a paired image. In total, the overall contrastive loss $\mathcal{L}_c$
\begin{align}
    \mathcal{L}_{c} = \alpha \mathcal{L}^{I2T}_c + (1-\alpha)\mathcal{L}^{T2I}_c,
\end{align}
where hyper-parameter $\alpha$ is the loss weighting coefficient.
\subsubsection{Cross-modal Knowledge Distillation}
% Goal:
To prevent catastrophic forgetting, knowledge distillation is introduced to keep the instance-wise prediction between current model $\mathcal{H}_t$ and the model $\mathcal{H}_{t-1}$ learned at last task.
Concretely, we retain an image's relationships with texts by employing the knowledge distillation loss $\mathcal{L}^{I2T}_d$ with KL-divergence as follows
\begin{align}
\begin{split}
\mathcal{L}^{I2T}_d = \frac{1}{B} \sum_{i=1}^{B} \text{KL}\bigg( P_{I2T}(y_i|\bm{\tilde{u}}_i, \{\bm{\tilde{v}}_j\}_{j=1}^{B_T};\theta_t, \tau ) ||  \\
 P_{I2T}(y_i|\bm{\tilde{u}}'_i, \{\bm{\tilde{v}}'_j\}_{j=1}^{B_T};\theta_{t-1}, \tau^d_{old}) \bigg),
\end{split}
\end{align}
where $\bm{\tilde{u}}'_i$,$\bm{\tilde{v}}'_j$ are the features extracted the model $\mathcal{H}_{t-1}$, $\tau^d_{old}$ represents the temperatures of last model for distillation.
Similarly, for a text $\bm{x}^T_j$, $1\le j\le B$, we also apply distillation loss on the prediction probabilities from text to image as follows
\begin{align}
\begin{split}
\mathcal{L}^{T2I}_d = \frac{1}{B} \sum_{j=1}^B \text{KL}\bigg(P_{T2I}(y_j|\bm{\tilde{u}}_j, \{\bm{\tilde{v}}_i\}_{i=1}^B;\theta_t, \tau) || \\ P_{T2I}(y_j|\bm{\tilde{u}}'_j, \{\bm{\tilde{v}}'_i\}_{i=1}^B;\theta_{t-1}, \tau^d_{old})\bigg).
\end{split}
\end{align}
Therefore, the knowledge distillation loss
\begin{align}
\mathcal{L}_d = \eta \mathcal{L}^{I2T}_d + (1-\eta)\mathcal{L}^{T2I}_d,
\end{align}
where the hyper-parameter $\eta$ is the loss weight.
% Why adopt knowledge distillation

\subsubsection{Overall Loss} Finally, we combine the contrastive loss $\mathcal{L}_c$ and distillation loss $\mathcal{L}_d$, and obtain  the final loss as follows
\begin{align}
    \mathcal{L}_{\text{overall}} = \mathcal{L}_{\text{c}} + \lambda \mathcal{L}_{\text{d}},
\end{align}
where $\lambda$ is the loss coefficient to control the tradeoff between losses.

Notably, we find that the weight norm often increases over different steps, which hampers the generalization ability of the pretrained model and causes negative forward transfer.
The same phenomenon is also observed in recent works~\cite{ash2020warm}.
Empirically, we adopt the trick of weight norm clipping. 
Concretely, at the end of each training iteration, if the weight norm at layer-l is higher than $\delta_l$, we clip the weight norm to $\delta_l$ when keeping the direction of the weight $W_l$ unchanged.
In practice, $\delta_l=\gamma \Vert W \rVert_{\text{init}}$ where the initial weight norm are denoted as $\Vert W \rVert_{\text{init}}$.
After the training of step $t$, we follow the practices~\cite{aljundi2019online,chaudhry2019tiny}  to update the memory by adopting reservoir sampling to select samples from the avaiable data to save.
%  randomly selecting examples from the available data.
% TODO:\xh{may need to add one paragraph "discussion": the additional computational cost, the storage usage benefit, the training speed/pseudo label generation speed}
% \paragraph{Negative Text Augmentation}
% We take the generated texts $\{e^{nT}_k\}$ as negative text to help the multimodal representation learning.
% For each training iteration, we sample a mini-batch pseudo text $\{e^{nT}_k\}$ with batch size $B^e_aug$
% The total batch size $B_{\text{aug}}$
% The set of texts $\{x^T\}_j=1^B\cup \{x^{nT}_k\}$ with $B_{\text{aug}}$ texts.
% Concretely, the probability from image to text now becomes
% \begin{align}
%     P'_{I2T}(y|x^I_i, \{x^T_j\}_{n=1}^{B_{\text{aug}};\tau) = \frac{e^{s_{in}/\tau}}{\sum_{n=1}^{B_{\text{aug}}}} e^{s_{in}/\tau}}.
% \end{align}
% Correspondingly, the contrastive loss from image to text now becomes
% \begin{align}
%     \mathcal{L}^{I2T}_{\text{caug}} = \frac{1}{B} \sum_{i=1}^B \log P'_{I2T}(y|x^I_i, \{x^T_j\}_{j=1}^B).
% \end{align}
% Moreover, the distillation loss from image to text becomes
% \begin{align}
% \begin{split}
% \mathcal{L}^{I2T}_{\text{daug}} = \frac{1}{B} \sum_{i=1}^B \text{KL}(P'_{I2T}(y|x^I_i, \{x^T_j\}_{j=1}^B;\theta_t, \tau) || \\
% P'_{I2 T}(y|x^I_i, \{x^T_j\}_{j=1}^B;\theta_{t-1}), \tau_{old})
% \end{split}
% \end{align}

%% file: data/exps.tex
\section{Experiments}\label{sec:exps}
In this section, we conduct exhaustive experiments to validate the effectiveness of our method.
Concretely, we first describe the experimental setup and implementation details in Sec.~\ref{subsec:exp_setup}, followed by the evaluation results on class incremental split in Sec.~\ref{subsec:cis}.
Then we introduce the experimental results on the instance incremental split in Sec.~\ref{subsec:iis}.
Finally, we perform ablation study and analysis to validate the effectiveness of components and provide more insights in Sec.~\ref{subsec:ablation}.

\begin{table*}[t]
\caption{\textbf{Results on class incremental CC2M dataset at final step}: The top-1 accuracy over various downstream datasets on zero-shot image classification task.}
\centering
% \normalsize
% \resizebox{0.83\textwidth}{!}{
% \begin{tabular}{ll|llllllll|l}
\begin{tabular}{ll|p{9mm}p{9mm}p{9mm}p{9mm}p{9mm}p{9mm}p{9mm}p{9mm}|p{9mm}}
\toprule
\#Tasks & Methods & \rotatebox{90}{ImageNet} & \rotatebox{90}{CIFAR-10} & \rotatebox{90}{CIFAR-100} & \rotatebox{90}{Caltech101}  & \rotatebox{90}{SUN397} & \rotatebox{90}{Food101} & \rotatebox{90}{Flowers102} & \rotatebox{90}{DTD} & \rotatebox{90}{Average} \\
\midrule
& Joint & 29.97 & 51.94 & 26.04 & 65.2 & 31.79 & 23.71 & 19.44 & 12.82 & 32.61 \\
\midrule
\multirow{6}{*}{4}&ER~\cite{chaudhry2019tiny} & 14.45 & 23.56 & 7.84 & 35.93 & 16.83 & 12.59 & 10.17 & 8.94 & 16.29  \\
% & WarmStart~\cite{} & & & & & & & & &  \\
& UCIR~\cite{hou2019learning} & 13.24 & 25.56 & 8.47 & 35.14 & 17.13 & 13.05 & 9.97 & \textbf{9.84} & 16.55 \\
% &GeoDL~\cite{simonKH21} &  & & & & & & & &  \\
& $\text{Co}^2\text{L}$~\cite{cha2021co2l} & 14.73 & 26.46 & 10.51 & 34.58 & 17.54 & 12.16 & 11.41 & 7.3 & 16.84 \\
& GeoDL~\cite{simon2021learning} & 14.24 & 27.48 & 9.49 & 35.93 & 17.01 & 12.94 & 9.87 & 8.99 & 17.00 \\
& IncCLIP & \textbf{18.85}& \textbf{28.31} & \textbf{13.23} & \textbf{50.32} & \textbf{23.38} & \textbf{16.19} & \textbf{13.08} & 9.20 & \textbf{21.57} \\
\midrule
\multirow{6}{*}{8}&ER~\cite{chaudhry2019tiny} & 9.59 & 14.23 & 3.85 & 26.80 & 11.95 & 7.24 & 8.98 & 5.48 & 11.02 \\
% & WarmStart~\cite{} & & & & & & & & &  \\
& UCIR~\cite{hou2019learning} & 9.89 & 12.69 & 4.42 & 23.56 & 12.95 & 9.53 & 9.33 & 7.02 & 11.17 \\
% & GeoDL~\cite{} & & & & & & & & &  \\
& $\text{Co}^2\text{L}$~\cite{cha2021co2l} & 10.99 & 18.34 & 5.51 & 29.1 & 13.52 & 9.01 & 8.56 & 5.9 & 12.62 \\
& GeoDL~\cite{simon2021learning} & 10.86 & 14.88 & 5.11 & 30.56 & 14.2 & 10.17 & 8.09 & 7.18 & 12.64 \\ 
% & Ours & \textbf{13.84} & \textbf{22.43} & \textbf{9.99} & \textbf{39.6} & \textbf{19.28} & \textbf{12.39} & \textbf{11.3} &  \textbf{7.43} & \textbf{17.03} \\
& IncCLIP & \textbf{13.93} & \textbf{22.68} & \textbf{10.45} & \textbf{43.39} & \textbf{19.27} & \textbf{11.91} & \textbf{9.57} &  \textbf{7.38} & \textbf{17.24} \\

\bottomrule
\end{tabular}%}
\label{tab:zs_cls}
\end{table*}

\begin{table*}[t]
\caption{\textbf{Image-text Retrieval Performance at final step:} Zero-shot Image-Text Retrieval on MSCOCO and Flickr30k datasets with various methods. R@K means top-K recall.}
\centering
\resizebox{\textwidth}{!}{
% \begin{tabular}{llllllllllllll}
\begin{tabular}{llp{8.8mm}p{8.8mm}p{8.8mm}p{8.8mm}p{8.8mm}p{8.8mm}p{8.8mm}p{8.8mm}p{8.8mm}p{8.8mm}p{8.8mm}p{8.8mm}}
\toprule
\multirow{3}{*}{\#Tasks} & \multirow{3}{*}{Methods} & \multicolumn{6}{c}{Flickr30K} & \multicolumn{6}{c}{MSCOCO} \\
& & \multicolumn{3}{c}{image-to-text} & \multicolumn{3}{c}{text-to-image} & \multicolumn{3}{c}{image-to-text} & \multicolumn{3}{c}{text-to-image} \\
&&  R@1 & R@5 & R@10 & R@1 & R@5 & R@10 & R@1 & R@5 & R@10 & R@1 & R@5 & R@10 \\
\midrule
& Joint & 35.7 & 62.4 & 71.8 & 25.16 & 50.52 & 62.36 & 18.8 & 41.82 & 52.94 & 13.14 & 30.74 & 41.78 \\
\midrule
\multirow{5}{*}{4} & ER~\cite{chaudhry2019tiny} & 17.5 & 39.8 & 51.3 & 11.18 & 28.68 & 38.18 & 9.6 & 24.72 & 35.34 & 6.41 & 17.76 & 25.61 \\
% & WarmStart & & & & & & & & & & & & \\
& GeoDL~\cite{simon2021learning} & 18.1 &  40.4& 51.1 & 11.96 & 29.94 & 39.4 & 9.64 & 25.58 & 35.78 & 6.26 & 18.02 & 26.01 \\
& UCIR~\cite{hou2019learning} & 18.3 & 41.9 & 52.9 & 12.10 & 30.32 & 40.28 & 9.72 & 25.88 & 36.16 & 6.64 & 18.58 & 26.83 \\
% & GeoDL~\cite{simonKH21} & & & & & & & & & & & \\
& Co$^2$L~\cite{cha2021co2l} & 19.7 & 42.5 & 53.1 & 12.24 & 29.34 & 39.54  & 10.1 & 25.16 & 35.24 & 6.78 & 18.30 & 26.59 \\
& IncCLIP & \textbf{24.1} & \textbf{49.5} &  \textbf{61.9} & \textbf{17.14} & \textbf{37.96} & \textbf{48.96} & \textbf{12.38} & \textbf{29.96} & \textbf{40.6} & \textbf{8.49} & \textbf{22.55} & \textbf{31.90} \\
\midrule
\multirow{5}{*}{8} & ER~\cite{chaudhry2019tiny} & 9.7 & 27.3 & 38.5 & 6.54 & 19.16 & 27.7 & 6.47 & 16.84 & 24.48 & 4.22 & 13.01 & 19.05  \\
% & WarmStart & & & & & & & & & & & & \\
& GeoDL~\cite{simon2021learning} & 12.5 & 33.3 & 42.1 & 8.40 & 21.9 & 30.44 & 6.42 & 18.08 & 27.32 & 4.64 & 14.02 & 20.58  \\
& UCIR~\cite{hou2019learning} & 10.1 & 28.7 & 40.1 & 7.16 & 20.54 & 29.14 & 7.00 & 17.58 & 25.22 & 4.38 & 13.13 & 19.83  \\
% & GeoDL~\cite{simonKH21} & & & & & & & & & & & \\
& Co$^2$L~\cite{cha2021co2l} & 12.9 & 32.3 & 41.9 & 8.43 & 21.76 & 30.41 & 6.22 & 18.68 & 26.58 & 4.49 & 13.29 & 20.24  \\
& IncCLIP & \textbf{16.0} & \textbf{37.7} &  \textbf{49.2} & \textbf{10.92} & \textbf{28.26} & \textbf{38.60} & \textbf{9.52} & \textbf{24.18} & \textbf{33.60} & \textbf{6.29} & \textbf{17.41} & \textbf{26.08} \\
\bottomrule
\end{tabular}}
\label{tab:i2t_retrieval}
	
\end{table*}

\begin{figure*}[t]
  \centering
%   \fbox{\rule{0pt}{2in} \rule{0.9\linewidth}{0pt}}
   \includegraphics[width=\linewidth]{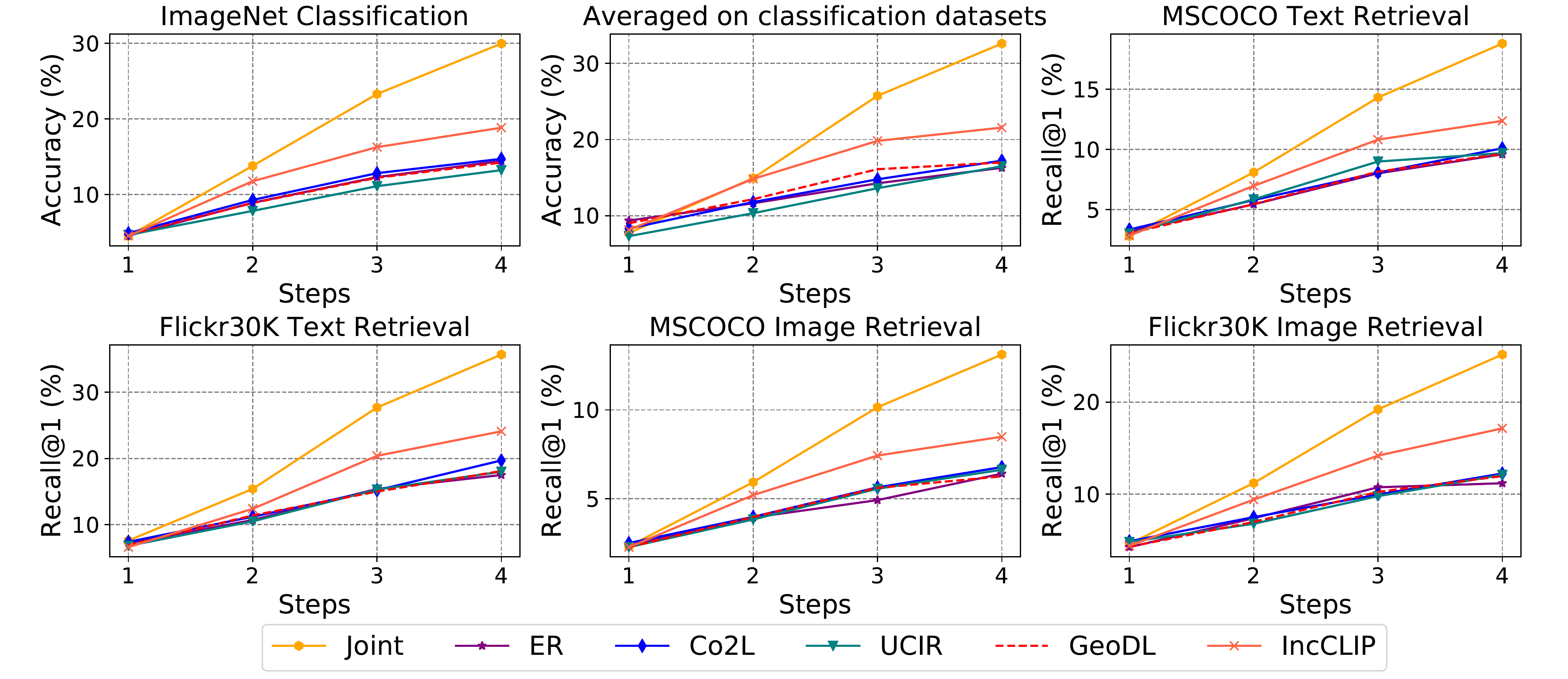} \\
   \caption{The downstream task performance over time. For image classification task, the accuracy is determined by averaging accuracies of all downstream datasets.}
   \label{fig:perf_wrt_steps}
\end{figure*}

\subsection{Experiment Setup and Implementation Details}\label{subsec:exp_setup}

\subsubsection{Benchmark Protocol} 
% Conceptual 12M (CC12M)~\cite{sharma2018conceptual} is a dataset collected from Internet including 12 million image-text pairs for vision-language pre-training.
% We adopt the CC12M dataset for two settings: \textit{(1) CC2M:}  two million image-text pairs are selected for training, and the data is randomly separated in four or eight chunks; \textit{(2) CC12M:}  all 12 million image-text pairs in CC12M dataset are  used, and the data is randomly divided into four chunks.
% We allow the algorithm to save fixed size of examples as memory during continual learning.
% We assess the quality of vision-language representation on downstream tasks including zero-shot image classification and zero-shot image-text retrieval~\cite{radford2021learning}.
Conceptual 12M (CC12M)~\cite{sharma2018conceptual} is a dataset collected from the Internet including 12 million image-text pairs for vision-language pre-training. 
In this section, we show results on both the class incremental and instance incremental split, corresponding to large and insignificant distribution shift in real world, respectively.
Considering that the image labels are not provided, we adopt an approximate strategy to build our class incremental split. Specifically, we first generate a pseudo class label for each image and then
partition ImageNet1k dataset into four chunks with 250 classes per chunk.
Here the pseudo labels are the most confident predictions
from the BEiT~\cite{bao2021beit} model pre-trained on ImageNet1k.
% To build class incremental split,  we approximate the class distribution by estimating image label with ImageNet1k pretrained model BEiT~\cite{bao2021beit}.
% With the pretrained model, we allocate the image label with the class with maximal prediction confidence, select 2M image-text pairs spanning all 1k classes, and then approximate the class distribution shift by partitioning them into 4 chunks with 250 classes per chunk.
For the instance incremental split, it models a real-world scenario in which data chunks are continuously collected in the same environment~\cite{hu2021well}.
In particular, we build the split via randomly selecting 2M image-text pairs from CC12M and then randomly split them into 4 identically distributed chunks with 0.5M image-text pairs each chunk.
% firstly we select two million image-text pairs from CC12M as CC2M dataset, and then build the class incremental split by their estimated image label to approximate the distribution shift of the representative real scenario.
%, where the incoming image-text pairs teaches the model novel concepts.
Note that the 2M image-text pairs used for instance incremental setting here is different with image-text pairs of the class incremental split for verifying the robustness of algorithm.
For completeness, we provide experiments for instance incremental split on same 2M image-text pairs with class incremental split in appendix.
The splits will be released in the future.
We allow the algorithms to store fixed-size instances in memory.

\begin{table*}[t]
\caption{\textbf{Ablation Study:} W.N.C means the weight norm cliping, H.N.T.G means hard negative text generation and Dist. means the knowledge distillation loss on image classification and image-to-text retrieval task. Below results on image-to-text retrieval tasks are top-1 recall.}
\centering
% \begin{tabular}{ccc|ccccc}
% \begin{tabular}{p{15mm}p{15mm}p{15mm}|p{16mm}p{12mm}p{12mm}p{12mm}p{12mm}}
\begin{tabular}{p{13mm}p{13mm}p{13mm}|p{16mm}p{12mm}p{12mm}p{12mm}p{12mm}}
\toprule
\multirow{2}{*}{W.N.C} & \multirow{2}{*}{Dist.} & \multirow{2}{*}{H.N.T.G} & \multirow{2}{*}{ImageNet} & \multicolumn{2}{c}{Flickr30K} & \multicolumn{2}{c}{MSCOCO}\\
& & & & I2T & T2I & I2T & T2I \\
\midrule
\checkmark & \xmark & \xmark & 16.01 & 21.3 & 14.06 & 10.89 & 7.52 \\
\checkmark& \checkmark & \xmark & 17.67 &  22.8 & 15.10 & 11.55 & 8.02 \\
\checkmark & \checkmark & \checkmark & \textbf{18.85} & \textbf{24.1} & \textbf{17.14} & \textbf{12.38} & \textbf{8.49}  \\
\bottomrule
\end{tabular}
\label{tab:ablation}
\end{table*}

% \begin{table*}[t]
% \caption{The sensitive study on memory size for image classification and image-to-text retrieval task.}
% \centering
% % \resizebox{0.8\textwidth}{!}{
% \begin{tabular}{ccccccccccccc}
% \toprule
%  & \multicolumn{4}{c}{ImageNet} & \multicolumn{8}{c}{MSCOCO} \\ 
%  &   &  & &  & \multicolumn{4}{c}{image-to-text} & \multicolumn{4}{c}{text-to-image} \\
%   & 1$\%$ & 5$\%$ & 10$\%$  & 20$\%$ & 1$\%$ & 5$\%$ & 10$\%$ & 20$\%$ & 1$\%$ & 5$\%$ & 10$\%$ & 20$\%$ \\
% \midrule
% IncCLIP & 14.05 & 14.69 & 15.10 & 16.34 & 12.92 & 12.28 & 13.46 & 14.62 & 7.88 & 8.28 & 8.60 & 9.70 \\
% \bottomrule
% \end{tabular}
% \label{tab:mem_size}
% \end{table*}

\begin{table*}[t]
\caption{The sensitive study of memory size for image classification task. `Average' means the average accuracy over all downstream image classification datasets.}
\centering
% \resizebox{0.8\textwidth}{!}{
\begin{tabular}{c|p{8mm}p{8mm}p{8mm}p{8mm}p{8mm}|p{8mm}p{8mm}p{8mm}p{8mm}p{8mm}}
\toprule
 & \multicolumn{5}{c|}{ImageNet} & \multicolumn{5}{c}{Average} \\ 
  & 1$\%$ & 5$\%$ & 10$\%$  & 20$\%$ & 50$\%$ & 1$\%$ & 5$\%$ & 10$\%$ & 20$\%$ & 50$\%$ \\
\midrule
IncCLIP & 12.82 & 16.68 & 18.95 & 23.61 & 28.29 & 16.90 & 20.16 & 21.82 & 24.92 & 30.61 \\
\bottomrule
\end{tabular}
\label{tab:mem_size_image}
\end{table*}

\begin{table*}[t]
\caption{The sensitive study on memory size for image-to-text retrieval on Flickr30K.}
\centering
% \resizebox{0.8\textwidth}{!}{
\begin{tabular}{c|p{8mm}p{8mm}p{8mm}p{8mm}p{8mm}|p{8mm}p{8mm}p{8mm}p{8mm}p{8mm}}
%ccccccccc}
\toprule
%  & \multicolumn{10}{c}{Flickr30K} \\ 
  & \multicolumn{5}{c|}{image-to-text} & \multicolumn{5}{c}{text-to-image} \\
  & 1$\%$ & 5$\%$ & 10$\%$  & 20$\%$ & 50$\%$ & 1$\%$ & 5$\%$ & 10$\%$ & 20$\%$ & $50\%$ \\
\midrule
IncCLIP & 18.2 & 21.6 & 24.1  & 27.2 & 31.6 & 13.16 & 15.58 & 17.14 & 18.38 & 23.62 \\
% Ours & 14.05 & 14.69 & 15.10 & 16.34 & 12.92 & 12.28 & 13.46 & 14.62 & 7.88 & 8.28 & 8.60 & 9.70 \\
\bottomrule
\end{tabular}
\label{tab:mem_size_text}
\end{table*}

% \paragraph{Comparison Methods}
\subsubsection{Comparison Methods}
To demonstrate the efficacy of our method, we adopt ER~\cite{chaudhry2019tiny}, UCIR~\cite{hou2019learning}, GeoDL~\cite{simon2021learning} as comparison methods.
% Experience replay(ER)~\cite{chaudhry2019tiny} refers that the model is finetuned on both memory and incoming data.
% UCIR~\cite{hou2019learning} is the classical distillation-based method in class-incremental learning. $\text{Co}^2L$~\cite{cha2021co2l} applies self-supervised distillation for self-supervised continual learning on images.
% GeoDL~\cite{simon2021learning} minimize the discrepancy between models by adopting geodesic distance connecting the low-dimensional manifolds.
Notably, we replace the cross entropy loss with contrastive loss to train a VLP model for ER, UCIR and GeoDL.
Besides, we also provide the joint-training(Joint) as an upper bound which makes use of all observed data to update the model.

\subsubsection{Implementation Details}
% Backbone
Following CLIP~\cite{radford2021learning}, all methods including comparison methods and our method IncCLIP adopt Transformer~\cite{vaswani2017attention} as the text backbone with the architecture modifications described in~\cite{radford2019language}.
Particularly, we use 12-layer 512-wide transformer with 8 attention heads and modified ResNet-50~\cite{radford2021learning}.
%Input, Augmentation
% The resolution of input images is resized to 224x224 for both pre-training on training set and evaluation on downstream tasks.
% The maximum sequence length of the tokens is limited to 77.
% We only use the center crop for data augmentation in the training.
% Prompt Template
The results on zero-shot image classification are reported using the prompt ensemble technique~\cite{radford2021learning}.
%Hyper-parameters
For each mini-batch, the size of sampled negative text $B_{\text{aug}}=256$.
The loss coefficient $\lambda=5$.
% The minimal and maximal scores $s_{\text{min}}=0$, $s_{\text{max}}=0.7$ used in the margin loss, and the threshold $\omega=0.05$.
$\alpha=0.5$, $\eta=0.5$, $\tau^d_{old}=0.01$.
% For the text generation, it is worth noting that when the loss $\mathcal{L}_G$ is smaller than $\omega=0.05$, we perform early stopping to accelerate the training.
% For the weight norm clipping, we set $\gamma$ to 0.7 on CC2M dataset. %and 1 on CC12M dataset.
The following experiments are conducted with $10\%$ dataset as memory except otherwise stated.
% Concretely, if not specified, all comparison methods save 0.2M image-text pairs for CC2M dataset.%, and 1.2M image-text pairs for CC12M dataset.
% Optimizer(To do)
To train our model, we adopt 8 Nvidia V100 GPUs with batch size 512 per GPU.
For each step, we train the model for 15 epochs on CC2M.% and 30 epochs on CC12M datasets.
We adopt LAMB optimizer with learning rate 0.003 and weight decay 0.003.
We begin by performing a linear warmup at each incremental step and then decay it using a cosine learning rate schedule.
More details can be found in appendix.

%Appendix:
% In the appendix, we detail the prompt templates for each downstream dataset.
% Concretely, we ensemble prompt templates by using their mean textual representation to calculate image-text similarity.

\subsection{Class Incremental Split}\label{subsec:cis}

\subsubsection{Zero-shot Image Classification}\label{subsec:zs_cls}
We conduct experiments to evaluate the algorithm's zero-shot generalization ability on image classification task.
Concretely, we evaluate models on eight representative datasets including CIFAR-10~\cite{krizhevsky2009learning}, CIFAR-100~\cite{krizhevsky2009learning}, Caltech101~\cite{fei2006one}, Oxford 102 Flower(Flowers102)~\cite{nilsback2008automated}, Food101~\cite{bossard14}, SUN397~\cite{barriuso2012notes}, Describable Textures Dataset (DTD)~\cite{cimpoi2014describing},  and ImageNet~\cite{deng2009imagenet}.
Like CLIP~\cite{radford2021learning}, we adopt prompt template, embed the class name to acquire the prediction score for each class, then use the class with the highest score as the prediction label during inference.

Table~\ref{tab:zs_cls} summarizes the final step performance of CC2M dataset with 4 and 8 steps.
We can see that our method regularly surpasses other methods with a significant margin at different downstream datasets.
Specifically, our method improves the accuracy from $19.7$ to $24.1(+\textbf{4.4}\%)$ under the 4 step split on ImageNet.
It is observed that when the number of steps increases, the average gain from our method increases as well.
In spite of the fact that traditional methods like UCIR,Co$^2$L are better than ER , it can be seen that the improvement is limited  compared to ER, indicating that their direct application can not fix the problem well.
Moreover, although our method achieves obvious gain, the large gap between our method and Joint(Upper bound) indicates that continual vision-language pre-training is challenging to be solved.
Furthermore, despite the clear increase achieved by our method, the wide gap between our method and the upper bound 'Joint' suggests that continuous vision-language pre-training is far from to be solved.
As shown in the Figure~\ref{fig:perf_wrt_steps}, it is observed that our method consistently outperforms the comparison methods at different steps.

% As Table~\ref{tab:cc12m_results} shows, our method constantly performs better than ER on CC12M dataset, indicating the superiority of our method.
% It is worth noting that the performance gain compared with ER decreases compared to the one in CC2M dataset with 4 steps.

% Length increases, the gap increases.

% \begin{table*}[t]
% \caption{CC12M: Zero-shot Top-1 Image Classification}
% \centering
% \resizebox{\textwidth}{!}{
% \begin{tabular}{l|llllllll|l}
% \toprule
% & \rotatebox{90}{\textbf{ImageNet}} & \rotatebox{90}{CIFAR-10} & \rotatebox{90}{CIFAR-100} & \rotatebox{90}{Caltech101} & \rotatebox{90}{SUN397} & \rotatebox{90}{Food101} & \rotatebox{90}{Flowers102} & \rotatebox{90}{DTD} & \rotatebox{90}{Average} \\
% \midrule
% Bound & & & & & & & & &  \\
% ER & & & & & & & & & \\
% WarmStart & & & & & & & & &  \\
% UCIR & & & & & & & & &  \\
% GeoDL & & & & & & & & & \\
% Ours & & & & & & & & & \\
% \bottomrule
% \end{tabular}}
% \end{table*}

% \vspace{-2mm}
\subsubsection{Zero-shot Image-Text Retrieval}\label{subsec:zs_itr}
The image-text retrieval task consists of two sub-tasks: image-to-text retrieval and text-to-image retrieval.
In particular, we employ MSCOCO~\cite{lin2014microsoft} and Flickr30K~\cite{plummer2017flickr30k} datasets to assess the representation transferability of pretrained representation on image-text retrieval task.
Table~\ref{tab:i2t_retrieval} demonstrates the image-text retrieval  results on 4-step split of CC2M. 
We can see that our method consistently outperforms the other methods in both image-to-text and text-to-image retrieval tasks.
Particularly, our method improves from $10.1\%$ to $12.38\%$ for top-1 text recall on MSCOCO dataset at final step.
As the number of steps increases from 4 to 8, our method's performance on all metrics falls, indicating that the continuous vision-language pre-training task becomes more difficult for longer sequences.
% \vspace{-2mm}
\subsection{Instance Incremental Split}\label{subsec:iis}
As Table~\ref{tab:iid_split} shows, we evaluate the methods on instance incremental split, and our method consistently outperforms than other methods on both classification and retrieval tasks tasks, showing the superiority and the robustness of our method. Our method achieves $2.12\%$ improvement on ImageNet classification task.
% \vspace{-2mm}
\subsection{Ablation Study and Analysis}\label{subsec:ablation}
\subsubsection{Ablation Study} 
We conduct exhaustive ablation study to evaluate the influence of each component used in our method.
As shown in Table~\ref{tab:ablation}, we can see that the generalization performance is improved when weight norm clipping is applied. 
Moreover, it is observed that the introduction of knowledge distillation loss gains $1.66\%$ improvement on classification accuracy on ImageNet.
Finally, with the addition of negative text replay, we can further obtain $1.3\%$ top-1 text recall performance gain on Flickr30K dataset for text retrieval task.

% \bein{table}[t]
% \caption{}
% \centering
% \begin{tabular}{cccccc}
% \toprule
% M
%
% \bottomrule
% \end{tabular}
% 
% \end{table}
\begin{table*}[t]
\caption{\textbf{Results on instance incremental split of 2M image-text pairs at final step:} `Average' means the accuracy averaged on eight classification datasets. Moreover, we report the top-1 recall on both MSCOCO and Flickr30K dataset.}
\centering
\begin{tabular}{p{15mm}|p{15mm}p{15mm}|p{12mm}p{12mm}p{12mm}p{12mm}}
\toprule
\multirow{2}{*}{Methods}  & \multirow{2}{*}{ImageNet} & \multirow{2}{*}{Average} & \multicolumn{2}{c}{Flickr30K} & \multicolumn{2}{c}{MSCOCO}\\
& & & I2T & T2I  & I2T  & T2I \\
\midrule
Joint & 20.20 & 24.58 & 27.20 & 19.02 & 14.58 & 9.90\\
\midrule
ER~\cite{chaudhry2019tiny} & 10.74 & 14.38 & 16.3 & 10.74 & 8.50 & 5.49 \\
% GeoDL & & & & & & \\
UCIR~\cite{hou2019learning} & 10.57 & 14.91 & 16.7 & 11.34 & 9.14 & 6.11\\
GeoDL~\cite{simon2021learning} & 10.85 & 14.16 & 16.4 & 10.82 & 8.62 & 5.65 \\
Co$^2$L~\cite{cha2021co2l} & 11.12 & 15.33 & 18.3 & 10.86& 8.58& 5.69\\
IncCLIP & \textbf{13.24} & \textbf{17.97} & \textbf{26.5} & \textbf{17.18} & \textbf{13.46} & \textbf{8.60} \\
\bottomrule
\end{tabular}
% \vspace{-2mm}
\label{tab:iid_split}
\end{table*}

\begin{table}[t]
\caption{The forgetting analysis on various methods.}
\centering
\begin{tabular}{p{15mm}p{12mm}p{12mm}p{12mm}p{12mm}p{12mm}}
\toprule
Methods & ER & UCIR & Co$^2$L & GeoDL & IncCLIP \\
\midrule
BWT & -31.93 & -28.68 & -30.83 & -31.41 & -18.48 \\
\bottomrule
\end{tabular}
\label{tab:forget}
% \vspace{-3mm}
\end{table}
% \vspace{-3mm}
\subsubsection{Sensitive Study on Memory Size}
We conduct sensitive study on memory size and report the results in Table~\ref{tab:mem_size_image} and~\ref{tab:mem_size_text}.
%image classification and image-and-text retrieval
`\%x' indicates that we set the memory size to be $x$ percents of the total size of the dataset.
We can find that the performance on all tasks consistently improves as the memory size increases, which demonstrates the effectiveness of replay strategy once more.
% \vspace{-3mm}
\subsubsection{Forgetting Analysis}
Additionally, we conduct experiments to analyze the algorithm's resistance to forgetting.
To measure the forgetting in the continual vision-language pre-training, we define the backward transfer(BWT) as the accuracy changes on each training chunk.
Detailedly, $\text{BWT}=\frac{1}{N-1} \sum_{i=2}^N \frac{1}{i} \sum_{j=1}^i A^i_j - A^j_j$ where $A^i_j$ means the accuracy of step $i$ model $\mathcal{H}_i$ on chunk $j\le i$.
The model has forgotten part of knowledge it acquired in chunk j when $A^i_j - A^j_j \le 0$.
As Table~\ref{tab:forget} shows, we can see that our method is less prone to catastrophic forgetting, which indicates the superiority of our method.

%% file: data/conclusion.tex
\section{Conclusions}\label{sec:conclusion}
In this work, we develop a replay-based continual learning framework for vision-and-language pre-training with two main contributions.
First, we perform model inversion to generate hard negative texts in token embedding space conditioned on the available images, and then use them to augment training.
Second, we adopt contrastive learning as our pre-training objective and introduce the knowledge distillation on the similarity scores between images and texts.
We conduct extensive experiments on Conceptual Caption dataset, and show that our learning strategy outperforms previous methods. In future work,  efficiency of the text generation module may be further improved.

\noindent \textbf{Acknowledgements}\quad This work was supported by Shanghai Science and Technology Program 21010502700, and Shanghai Frontiers Science Center Program. This work was done when Shipeng Yan was intern at Noah’s Ark Lab. Hang Xu and Xuming He are equal corresponding authors. We gratefully acknowledge the support of MindSpore, CANN(Compute Architecture for 
Neural Networks) and Ascend AI Processor used for this research.